\pdfoutput=1
\documentclass[11pt,a4paper]{article}

\usepackage[utf8]{inputenc}
\usepackage[T1]{fontenc}
\usepackage{lmodern}
\usepackage{microtype}
\usepackage{geometry}
\usepackage{amsmath, amssymb, amsfonts}
\usepackage{mathtools}
\usepackage{bm}            
\usepackage{physics}       
\usepackage{enumitem}      
\usepackage{algorithm}
\usepackage{algpseudocode}
\setlist{topsep=4pt}

\usepackage{amsthm}

\theoremstyle{plain}

\theoremstyle{definition}

\theoremstyle{remark}

\usepackage[dvipsnames]{xcolor}
\usepackage[
   colorlinks = true,
   linkcolor = MidnightBlue,
   citecolor = MidnightBlue,
   urlcolor  = MidnightBlue
]{hyperref}
\usepackage[numbers]{natbib}
\usepackage{url}
\usepackage{cleveref}
\crefname{definition}{Definition}{Definitions}
\crefname{proposition}{Proposition}{Propositions}
\crefname{theorem}{Theorem}{Theorems}
\crefname{corollary}{Corollary}{Corollaries}
\crefname{lemma}{Lemma}{Lemmas}
\crefname{remark}{Remark}{Remarks}

\usepackage{graphicx}
\usepackage{caption}
\usepackage{subcaption}
\usepackage{booktabs}

\usepackage{tikz}
\usetikzlibrary{
   arrows.meta,
   decorations.pathmorphing,
   calc,
   positioning,
   shapes.geometric,
   decorations.markings,
   fit
}
\usepackage{pgfplots}
\pgfplotsset{compat=1.18}

\allowdisplaybreaks

\newcommand{\githuburl}{https://github.com/ShinobuMiya/Eidoku}
\begin{document}

\title{
Eidoku: A Neuro-Symbolic Verification Gate for LLM Reasoning via Structural Constraint Satisfaction
}

\author{
Shinobu Miya \\
Independent Researcher
}

\date{}

\maketitle

\begin{abstract}
Large Language Models (LLMs) frequently produce hallucinated statements that are assigned high likelihood by the model itself, exposing a fundamental limitation of probability-based verification.
This suggests that hallucination is often not a low-confidence phenomenon, but a failure of structural consistency.
In this work, we reformulate the verification of LLM reasoning as a \emph{Constraint Satisfaction Problem (CSP)} operating independently of the generation likelihood.
Rather than optimizing for statistical plausibility, we model verification as a feasibility check based on \emph{structural violation cost}---the computational cost required to embed a candidate reasoning step into the contextual graph structure.
We define a total cost function composed of three proxies: (i) graph connectivity (structural), (ii) feature space consistency (geometric), and (iii) logical entailment (symbolic).
Crucially, verification is performed via a lightweight System-2 gate, \textsc{Eidoku}, which rejects candidates exceeding a context-calibrated cost threshold.
The threshold is not learned but is derived from the intrinsic statistics of the context, avoiding ad hoc heuristics.
We demonstrate that this approach successfully rejects ``smooth falsehoods''---statements that are highly probable yet structurally disconnected---that probability-based verifiers are principally incapable of detecting.
Our experiments on a controlled diagnostic dataset show that explicitly enforcing structural constraints allows for the deterministic rejection of this specific class of hallucinations, serving as a neuro-symbolic sanity check for generative reasoning.

\vspace{0.5em}
\noindent
\textbf{Code Availability:} A reference implementation and reproduction logs are available at \url{\githuburl}.
\end{abstract}

\section{Introduction}

Large Language Models (LLMs) have demonstrated remarkable performance in natural language generation and multi-step reasoning.
However, they remain prone to producing hallucinated statements: fluent, confident outputs that are factually or logically incorrect.
Crucially, such hallucinations are often assigned high likelihood by the model itself, rendering probability-based confidence measures unreliable for verification.

This exposes a fundamental mismatch between generation and verification.
Likelihood, attention weights, and decoding probabilities are optimized for plausibility under the model distribution, not for semantic or logical feasibility within a given context.
As a result, high-probability outputs may still violate structural constraints, break transitive relations, or introduce unsupported entities, while appearing locally coherent.

Most existing approaches attempt to mitigate hallucinations by refining generation strategies, incorporating external knowledge, or estimating uncertainty from probabilistic signals \cite{huang2023survey}.\footnote{The source code for our Optimization-Independent verification gate, \textsc{Eidoku}, is open-sourced at \url{\githuburl}.}
\textsc{Eidoku} operates on an evaluation axis that is \textbf{optimization-independent} to the generation objective.
While System 1 optimizes for \emph{plausibility} (likelihood $P_\theta(y|x)$), System 2 evaluates \emph{structural feasibility} (score $J(S)$).
These two objectives are independent; a high-likelihood output can have high structural Violation Cost (low feasibility).
Thus, ``Independence'' here refers to the independence of the optimization landscapes, not necessarily the absence of probabilistic components in the scalar definitions of Violation Cost proxies.
In contrast, we argue that hallucination is better understood as a \emph{structural failure}: a candidate reasoning sequence cannot be embedded into the contextual semantic structure without incurring excessive deformation.

In this work, we propose to reformulate verification as an Optimization-Independent optimization problem over semantic structure.
Rather than estimating how probable an output is, we evaluate how much \emph{semantic Violation Cost} is required to accommodate it.
Semantic Violation Cost is defined as the cost of deforming contextual structure to integrate a candidate reasoning step.
Verification is then performed by minimizing accumulated Violation Cost under a calibrated critical threshold, enforcing closure of the semantic system.

We instantiate this formulation as \textsc{Eidoku}, a lightweight System-2 verification gate that operates independently of the generation process.
\textsc{Eidoku} evaluates candidate reasoning sequences produced by an LLM (System 1), rejects those that exceed a context-calibrated Violation Cost threshold, and selects the lowest-score accepted candidate.
A minimal proof-of-concept implementation demonstrates that \textsc{Eidoku} can reject trivial hallucinations that remain highly probable under standard decoding.

Our contributions are threefold:
(1) we propose an Optimization-Independent formulation of verification based on semantic structural Violation Cost;
(2) we introduce an additive score model composed of independent structural, geometric, and logical proxies;
and (3) we present a context-calibrated rejection mechanism that avoids fixed or learned thresholds.
Finally, we clarify our design scope: \textsc{Eidoku} does not attempt to model any underlying physical or semantic dynamics.
Instead, it adopts a pragmatic engineering assumption: treating the verification window as an isolated evaluation context functions as a regularization mechanism.

\paragraph{Note on Terminology:}
Throughout this work, we use the term \textbf{Optimization-Independent} to strictly denote independence of the \emph{optimization objective}, distinct from statistical independence of variables.
While internal components (e.g., NLI) may derive from probabilistic models, \textsc{Eidoku} utilizes them as scalar proxies for structural Violation Cost within a feasibility constraint ($J(S) \le \tau_c$), rather than for maximizing generative plausibility ($P(S) \to \max$).
Thus, ``Independence'' asserts that the verification axis is not aligned with the gradient of the generation likelihood.

\section{Why Likelihood Is \textbf{Insufficient} for \textbf{Structural} Verification}

Verification is commonly treated as a problem of confidence estimation.
Under this view, a model output is considered reliable if it is assigned high likelihood, low entropy, or strong attention alignment.
However, this assumption conflates statistical plausibility with semantic feasibility.

Likelihood reflects how well an output matches the model's learned distribution, not whether it is structurally compatible with the given context.
A language model may confidently produce a statement that is locally consistent in form yet globally incompatible with prior constraints.
For example, given a context encoding a transitive relation $A \rightarrow B \rightarrow C$, the statement ``$A$ is a fish'' may be assigned non-negligible probability if it matches common syntactic and semantic patterns, despite having no structural support in the context.

This failure mode is not incidental.
Probability-based decoding optimizes for continuation plausibility, not for global consistency.
Attention mechanisms, similarly, indicate relevance within the model's internal computation but do not constitute a semantic metric.
As such, neither likelihood nor attention \textbf{alone} provides a principled criterion for rejecting structurally unsupported reasoning steps.

Importantly, hallucinations are not necessarily low-probability events.
Empirically, many hallucinated statements are generated with high confidence and persist under temperature scaling or beam search.
This suggests that hallucination cannot be reliably detected by thresholding probabilistic scores alone.

We therefore argue that verification requires a different axis.
Rather than asking how likely a statement is, verification should ask whether a statement can be embedded into the contextual semantic structure without violating its constraints.
This reframes verification as a feasibility problem: does there exist a low-cost deformation of the semantic structure that accommodates the candidate?

Under this view, inconsistency manifests as excessive deformation cost.
A reasoning step that cannot be integrated without incurring high structural, geometric, or logical Violation Cost should be rejected, regardless of its likelihood.
This perspective motivates a verification mechanism that operates outside the probabilistic generation loop and evaluates candidates based on accumulated semantic Violation Cost.

The remainder of this paper develops this formulation and demonstrates its implications through a minimal System-2 verification gate.

\section{Geometric Cost Formulation}
\label{sec:formulation}

We formalize verification by defining a \textbf{Cost-Based Formulation} over the semantic structure of reasoning chains.
Instead of relying on probabilistic likelihoods, we map the reasoning process onto a geometric embedding space.

\textsc{Eidoku} instantiates this framework as follows:
\begin{itemize}
    \item The discrete vocabulary set $\mathcal{V}$ (tokens) serves as the state space.
    \item The embedding function $\phi: \mathcal{V} \to \mathbb{R}^d$ maps discrete tokens to a continuous semantic embedding space $M_{\phi} \subset \mathbb{R}^d$.
    \item The reasoning sequence $S = (S_1, \dots, S_n)$ represents a discrete trajectory on $M_{\phi}$.
\end{itemize}

We define the Violation Cost $\tau(S_i, S_j)$ not as a probabilistic surprisal, but as a \emph{constraint violation cost} required to connect steps $S_i$ and $S_j$ within the contextual constraints.
Specifically, we approximate the score gradient on the embedding space:
$$
\tau_{\text{Eidoku}}(S_i, S_j) \;\cong\;
\left\| \nabla J_{\text{embedding space}} \right\|_{M} \cdot \Delta S_{ij},
$$
instantiated via three independent proxies representing structural, geometric, and logical constraints.

Let the context be a finite set of prior statements, $\mathcal{C} = \{C_1, C_2, \dots, C_m\}$.
Our goal is to select the candidate $S^\ast$ that minimizes the total accumulated score:
$$
S^\ast = \arg\min_{S \in \mathcal{S}} J(S),
$$
where
$$
J(S) = \sum_{(i,j)\in E(S)} \tau(S_i, S_j \mid \mathcal{C}).
$$
In this formulation, boundary conditions (such as the introduction of entities not grounded in the context) are treated as structural discontinuities (infinite score barrier) within $\tau_{\text{struct}}$.

\subsection{Reasoning Graph and Accumulated Violation Cost}
\label{subsec:reasoning-graph}

A reasoning sequence induces a set of local semantic junctions (adjacency, reference, entailment links, etc.) between steps.
We represent these junctions as edges in a graph
$$
G(S) = (V, E(S)), \quad V = \{S_1,\dots,S_n\},
$$
where $(i,j)\in E(S)$ denotes that $S_i$ constrains $S_j$ under the intended reasoning structure (e.g., $j=i+1$ for adjacency, or a longer-range reference).
We define the Violation Cost sum as
$$
J(S) = \sum_{(i,j)\in E(S)} \tau(S_i, S_j \mid \mathcal{C}).
$$

\subsection{Total Violation Cost as an Additive Function of Independent Proxies}
\label{subsec:tau-decomposition}

We model semantic Violation Cost $\tau$ as the \emph{cost required to deform contextual structure} so that the transition from $S_i$ to $S_j$ becomes feasible.
We decompose this total score into three independent components, analogous to potential terms in physical systems:

\begin{itemize}
    \item $\tau_{\text{struct}}$ (Topology): Connectivity cost in the knowledge graph.
    \item $\tau_{\text{curv}}$ (Geometry): Geometric deviation from the local embedding space.
    \item $\tau_{\text{logic}}$ (Logic): Logical entailment divergence.
\end{itemize}

Assuming a linearized approximation ($p=1$), the total Violation Cost is:
$$
\tau(S_i,S_j \mid \mathcal{C}) \;=\;
\frac{1}{\sigma_L}\,\tau_{\text{struct}}
\;+\;
\frac{1}{\sigma_G}\,\tau_{\text{curv}}
\;+\;
\frac{1}{\sigma_I}\,\tau_{\text{logic}},
$$
where $w_k = 1/\sigma_k$ are \emph{dimensional normalization factors} derived from the empirical standard deviation of each proxy on a calibration set.
This normalization ensures that the contribution of each component is dimensionless and variance-balanced, removing the dependency on arbitrary hyperparameters.
$\tau_{\text{struct}}$ penalizes the need for ad hoc structural bridges,
$\tau_{\text{curv}}$ penalizes geometric deviation from the local semantic embedding space implied by the context,
and $\tau_{\text{logic}}$ penalizes failures of entailment.

\subsection{Theoretical Justification: embedding space Consistency}
\label{subsec:theoretical-justification}

The score minimization principle operationalized in \textsc{Eidoku} rests on established principles in graph theory and embedding space learning.
We formally define the mechanism by which score minimization suppresses hallucinations.

\paragraph{Structural Closure as Connectivity:}
In graph theory, a reasoning chain is valid only if it forms a connected path within the transitive closure of the knowledge graph $G_{\mathcal{C}}$.
Let $\mathcal{R}(G_{\mathcal{C}})$ be the reachability set of the context.
A structural hallucination corresponds to introducing a node $u \notin \mathcal{R}(G_{\mathcal{C}})$.
By defining $\tau_{\text{struct}}$ as the shortest-path distance (where disconnection implies $d=\infty$), minimizing $J(S)$ is mathematically equivalent to enforcing \emph{connectivity constraints} on the generated subgraph \cite{cormen2009introduction}.

\paragraph{Geometric Regularization as embedding space Smoothness:}
In embedding space learning, semantic consistency implies that valid inferences lie on a low-dimensional subspace embedded in the high-dimensional latent space \cite{tenenbaum2000global}.
A geometric hallucination is a point $S_j$ that deviates from the local tangent space of the context embedding space.
Minimizing $\tau_{\text{curv}}$ (reconstruction error) is equivalent to enforcing \emph{local linearity} (LLE) or smoothness constraints.
Thus, the context window acts as a regularization constraint that penalizes off-embedding space projections, preventing the model from traversing into regions weakly supported by the context statistics.

In summary, \textsc{Eidoku} translates the abstract notion of structural closure into concrete topological constraints: ensuring that the reasoning trajectory remains within the connected component (structure) and on the smooth surface (geometry) of the context.

\subsubsection{Structural Violation Cost \texorpdfstring{$\tau_{\text{struct}}$}{tau-struct}}
\label{subsubsec:tau-struct}

Let $\Gamma(\mathcal{C})$ be a lightweight contextual structure extracted from $\mathcal{C}$, such as an entity-relation graph constructed by dependency parsing or OpenIE-style triple extraction.
We interpret a candidate transition as structurally supported if there exists a short, low-complexity path connecting its key entities in $\Gamma(\mathcal{C})$.
Let $d_{\Gamma}(u,v)$ be the shortest-path length between entities $u$ and $v$ in $\Gamma(\mathcal{C})$ (or $\infty$ if disconnected).
Then a minimal proxy is
$$
\tau_{\text{struct}}(S_i,S_j \mid \mathcal{C})
=
\alpha_{\text{struct}}\,
\log\!\bigl(1 + d_{\Gamma}(u_i,u_j)\bigr)
\;+\;
\beta_{\text{struct}}\,\mathbb{I}[d_{\Gamma}(u_i,u_j)=\infty],
$$
where $u_i$ and $u_j$ are canonical entities extracted from $S_i,S_j$ (e.g., subject terms), and $\mathbb{I}[\cdot]$ is the indicator function.
Intuitively, if no contextual path exists, the model would need to invent an auxiliary bridge, corresponding to high structural deformation cost.

This proxy can be extended with an MDL-inspired term.
Let $\mathcal{G}_{\text{adhoc}}$ denote the minimal auxiliary subgraph needed to connect $u_i$ to $u_j$.
Then an alternative structural Violation Cost is
$$
\tau_{\text{struct}}(S_i,S_j \mid \mathcal{C})
=
\operatorname{DL}\!\bigl(\mathcal{G}_{\text{adhoc}}\bigr),
$$
where $\operatorname{DL}(\cdot)$ is a description length of the introduced structure (e.g., number of nodes/edges plus label complexity).
In both forms, the objective is to assign high cost to structurally unsupported transitions.
We acknowledge that robust graph extraction from natural language is an open challenge.
Crucially, however, extraction failures typically manifest as \emph{missing edges} (false disconnections) rather than hallucinatory edges.
This asymmetry implies that imperfect graph extraction degrades \emph{utility} (TTAR) but preserves \emph{safety} (FTAR).
Consequently, the system is inherently biased towards \emph{conservative rejection} (fail-safe), prioritizing the prevention of hallucinations over the recall of valid reasoning in high-stakes settings.

\subsubsection{Geometric Violation Cost \texorpdfstring{$\tau_{\text{curv}}$}{tau-curv}}
\label{subsubsec:tau-curv}

Let $\phi(\cdot)\in\mathbb{R}^d$ be a fixed text embedding function (e.g., sentence embeddings), treated as a local coordinate chart rather than meaning itself.
Given the context window $\mathcal{W}\subset\mathcal{C}$ relevant to the current step, define the context matrix
$$
X = \begin{bmatrix}
\phi(C_{a_1})^\top \\
\phi(C_{a_2})^\top \\
\vdots \\
\phi(C_{a_k})^\top
\end{bmatrix}\in\mathbb{R}^{k\times d}.
$$
We estimate a local tangent subspace by PCA/SVD.
Let $U_r\in\mathbb{R}^{d\times r}$ be the top-$r$ right singular vectors of the centered $X$.
For a candidate step $S_j$, define the residual (projection) error
$$
\epsilon_{\text{lle}}(S_j \mid \mathcal{C})
=
\left\|
\left(I - U_r U_r^\top\right)\bigl(\phi(S_j)-\mu\bigr)
\right\|_2,
$$
where $\mu$ is the mean context embedding.
We then set
$$
\tau_{\text{curv}}(S_j \mid \mathcal{C})
=
\alpha_{\text{curv}}\,\log\!\bigl(1+\epsilon_{\text{lle}}(S_j \mid \mathcal{C})\bigr).
$$
This term penalizes steps that deviate strongly from the locally reconstructed semantic embedding space, capturing a geometric notion of ``off-embedding space'' statements.

\subsubsection{Logical Violation Cost \texorpdfstring{$\tau_{\text{logic}}$}{tau-logic}}
\label{subsubsec:tau-logic}

Logical Violation Cost penalizes failures of entailment from prior steps and the context.
Let $\operatorname{NLI}(p,h)\in[0,1]^3$ return probabilities for \{entailment, neutral, contradiction\}.
Define an entailment score
$$
e_{i\rightarrow j} = \operatorname{NLI}_{\text{ent}}(S_i, S_j),
\qquad
c_{i\rightarrow j} = \operatorname{NLI}_{\text{con}}(S_i, S_j).
$$
A simple proxy is
$$
\tau_{\text{logic}}(S_i,S_j \mid \mathcal{C})
=
\alpha_{\text{logic}}\,(1 - e_{i\rightarrow j})
\;+\;
\beta_{\text{logic}}\,c_{i\rightarrow j}.
$$
\footnote{Note that while the NLI module outputs a probability score, within \textsc{Eidoku} this serves strictly as a scalar proxy for logical distance (potential score) within the Violation Cost function, distinct from the generative likelihood of the sequence itself.}
Thus, steps that are not entailed incur cost, while explicit contradictions incur larger cost.

\subsection{Context-Adaptive Distance via Mahalanobis Scaling}
\label{subsec:metric}

To account for the varying rigidity of semantic constraints, we introduce a context-adaptive distance metric.
Instead of treating the embedding space as isotropic (Euclidean), we model the local semantic logic using a statistical manifold derived from the context window.

Given the centered context matrix $X \in \mathbb{R}^{k\times d}$ (where rows are step embeddings), we consider the empirical covariance matrix $\Sigma = \frac{1}{k}X^\top X$.
The eigen-decomposition $\Sigma = Q \Lambda Q^\top$ identifies the principal directions of the current reasoning context.
Directions with large eigenvalues correspond to high variance (semantically flexible axes), whereas directions with near-zero eigenvalues correspond to invariant constraints (rigid axes).

We define a precision matrix $M$ (inverse covariance proxy) to reweight the feature space:
$$
M = Q\,\operatorname{diag}\!\left(\frac{1}{\lambda_1+\epsilon},\dots,\frac{1}{\lambda_d+\epsilon}\right)Q^\top,
$$
where $\epsilon>0$ is a regularization term for numerical stability.
This matrix $M$ essentially acts as a soft constraint enforcer: it assigns high cost to deviations along directions where the context was historically invariant.

Operationally, we replace the Euclidean residual norm defined in Section~\ref{subsubsec:tau-curv} with the \emph{Mahalanobis norm}:
$$
\epsilon_{\text{lle}}^M(S_j \mid \mathcal{C})
=
\sqrt{
\bigl(\phi(S_j)-\mu\bigr)^\top
M
\bigl(\phi(S_j)-\mu\bigr)
}.
$$
This transformation ensures that the geometric penalty is sensitive to the \emph{structure} of the context, not just raw distance. A deviation of unit length is penalized heavily if it violates a rigid contextual invariant, but tolerated if it aligns with a high-variance axis.

\subsection{Closure and Rejection Rule}
\label{subsec:closure}

To enforce a closed verification regime, we introduce a critical Violation Cost threshold $\tau_c$.
A candidate is rejected if any local junction exceeds $\tau_c$:
$$
\textsc{Reject}(S) \;\;\text{if}\;\; \exists (i,j)\in E(S)\;\;\text{s.t.}\;\;\tau(S_i,S_j \mid \mathcal{C}) > \tau_c.
$$
This captures the principle that structurally unsupported steps should not be accepted even if they appear plausible under the model distribution.

\subsection{Per-Context Calibration of \texorpdfstring{$\tau_c$}{tau\_c} Without Learning}
\label{subsec:tau-calibration}

A fixed global threshold can appear arbitrary.
Instead, we calibrate $\tau_c$ per context from the observed Violation Cost distribution, without learning a predictive model.
Let
$$
\mathcal{T}(\mathcal{C}) = \{\tau(S_i,S_j \mid \mathcal{C}) \;:\; (i,j)\in E(S),\; S\in\mathcal{S}\}
$$
denote the set of observed local Violation Costs across all candidate junctions under the context.
We define
$$
\tau_c(\mathcal{C})
=
\operatorname{Percentile}_{p}\bigl(\mathcal{T}(\mathcal{C})\bigr)\cdot (1+\delta),
$$
with typical choices $p=95$ and $\delta \in [0.05,0.2]$.
This yields a context-adaptive critical threshold that is neither learned nor purely heuristic: it is a statistical definition of the ``critical'' tail region of the observed Violation Cost landscape.

In summary, the proposed formulation treats verification as an score minimization problem under a closure constraint.
The next section describes a minimal implementation, \textsc{Eidoku}, that operationalizes this formulation as a lightweight System-2 gate.

\section{Threshold Calibration as Outlier Detection}
\label{sec:threshold}

A central design choice in the proposed formulation is the use of a critical cost threshold $\tau_c$ for rejection.
Rather than viewing $\tau_c$ as a heuristic hyperparameter, we interpret it through the lens of \emph{statistical outlier detection}.

\subsection{Threshold as the Feasibility Boundary}
\label{subsec:boundary}

The threshold $\tau_c$ defines the \emph{feasibility boundary} of the semantic region supported by the context.
Candidates exceeding $\tau_c$ effectively lie in sparse, undefined regions of the latent space (out-of-distribution).
Our statistical calibration serves as a non-parametric estimator of this boundary.

Intuitively, a context with low overall variance (tight logical constraints) implies a steep cost landscape, tolerating less error.
Conversely, a vague or high-variance context implies a flat landscape, allowing larger steps.
By calibrating $\tau_c$ based on the internal statistics of the context, \textsc{Eidoku} adapts its rejection sensitivity to the local ``sharpness'' of the reasoning problem.

\subsection{Why a Fixed Global Threshold Is Inadequate}

A single global threshold implicitly assumes that semantic rigidity is uniform across all contexts.
This assumption is unrealistic.
Different contexts exhibit different degrees of redundancy, constraint density, and structural tightness.
For example, a tightly constrained mathematical context should tolerate far less deformation than an open-ended narrative context.

Using a fixed $\tau_c$ therefore introduces two failure modes:
(1) over-rejection in flexible contexts, and
(2) under-rejection in rigid contexts.
Both undermine the goal of Optimization-Independent verification.

\subsection{Observed Violation Cost Distributions as Contextual Statistics}

Rather than treating $\tau_c$ as a learned parameter, we treat it as a \emph{contextual statistic}.
Given a fixed context $\mathcal{C}$ and a set of candidate reasoning sequences $\mathcal{S}$ produced by System~1, the verification process naturally observes a collection of local Violation Cost values,
$$
\mathcal{T}(\mathcal{C})
=
\{\tau(S_i,S_j \mid \mathcal{C}) \;|\; (i,j)\in E(S),\; S\in\mathcal{S}\}.
$$
This set defines an empirical Violation Cost landscape induced by the context.

Importantly, this distribution exists independently of any acceptance decision.
It is a byproduct of evaluating candidates, not a trained model or a tuned heuristic.

\subsection{Dual-Stage Thresholding with Absolute Safety Bounds}
\label{subsec:tau-calibration-dual}

We define the critical threshold $\tau_c$ via a dual-stage mechanism that imposes both adaptive and absolute constraints.
Formally, let $\tau_{\text{raw}} = \operatorname{Percentile}_{p}\bigl(\mathcal{T}(\mathcal{C})\bigr)\cdot(1+\delta)$.
The operational threshold is defined as:
$$
\tau_c(\mathcal{C})
\;=\;
\min\!\left(
    \tau_{\text{max}}, \;
    \max\!\left(
        \tau_{\text{min}}, \;
        \tau_{\text{raw}}
    \right)
\right).
$$
This formulation integrates two distinct layers of defense against hallucinations:

\begin{enumerate}
    \item \textbf{Adaptive Elasticity (The Core logic):}
    The term $\tau_{\text{raw}}$ adapts to the contextual rigidity. In loose contexts, the threshold naturally relaxes, identifying which candidate is \emph{relatively} most coherent.

    \item \textbf{Absolute Safety Ceiling ($\tau_{\text{max}}$):}
    This imposes a hard upper bound on permissible Violation Cost.
    In a \textbf{Mode Collapse} scenario where all candidates are hallucinations (i.e., $\forall S \in \mathcal{S}, \tau(S) > \tau_{\text{max}}$), the threshold clamps at $\tau_{\text{max}}$.
    Consequently, all candidates are rejected, resulting in a \emph{Null Set} (System-level Refusal). This prevents the system from being forced to choose the ``least bad'' falsehood.

    \item \textbf{Numerical Stability Floor ($\tau_{\text{min}}$):}
    The lower bound $\tau_{\text{min}}$ acts strictly as a numerical stability constraint.
    It prevents \emph{degenerate rejection} in scenarios where the context is fully coherent and noise-free (i.e., variance $\to 0$).
    Without this floor, the adaptive percentile could collapse to near-zero, causing the rejection of valid but non-zero score candidates due to floating-point sensitivity.
\end{enumerate}

Thus, $\tau_c$ is not a learned decision boundary, but a context-adaptive statistic bounded by the physical stability limits of the embedding space.
This definition has three important properties.
First, $\tau_c$ adapts automatically to the scale and spread of Violation Costs induced by the context.
Second, it does not require any labeled data or gradient-based optimization.
Third, it defines ``criticality'' in purely statistical terms: a candidate step is rejected if it lies in the extreme tail of the observed Violation Cost landscape.

\subsection{Relation to Closure}

The threshold $\tau_c$ enforces a closed verification regime.
Once a local transition exceeds $\tau_c$, the system refuses to accommodate it by inventing unbounded auxiliary structure.
This corresponds to rejecting semantic ``escape routes'' that would otherwise allow arbitrary hallucinations to be rationalized post hoc.

Unlike probabilistic confidence thresholds, which compare scores across different contexts or models, $\tau_c$ is meaningful only relative to the current context.
It therefore does not serve as a universal notion of correctness, but as a context-specific boundary of admissible deformation.

\subsection{Distinction from Learning-Based Calibration}

It is important to distinguish the proposed calibration from learning-based threshold selection.
No loss function is minimized, no ground-truth labels are required, and no parameters are updated across contexts.
The calibration depends solely on the observed distribution of Violation Costs under the current context.

This design choice is intentional.
Learning a global decision boundary would reintroduce probabilistic assumptions and entangle verification with model-specific artifacts.
By contrast, per-context calibration preserves the Optimization-Independent nature of the framework and maintains a clear separation between generation (System~1) and verification (System~2).

\subsection{Implications}

By defining $\tau_c$ as a contextual statistic rather than a tunable parameter, the verification problem is reframed as identifying structural outliers within a semantic Violation Cost landscape.
This avoids ``magic constants'' while retaining a crisp rejection criterion.
More broadly, it supports the view that hallucination is a context-relative structural failure, not a low-confidence event.

The next section describes how this calibration is instantiated in a minimal implementation, and how rejection emerges naturally from the observed Violation Cost distribution.

\section{Eidoku: A Minimal System-2 Verification Gate}
\label{sec:eidoku}

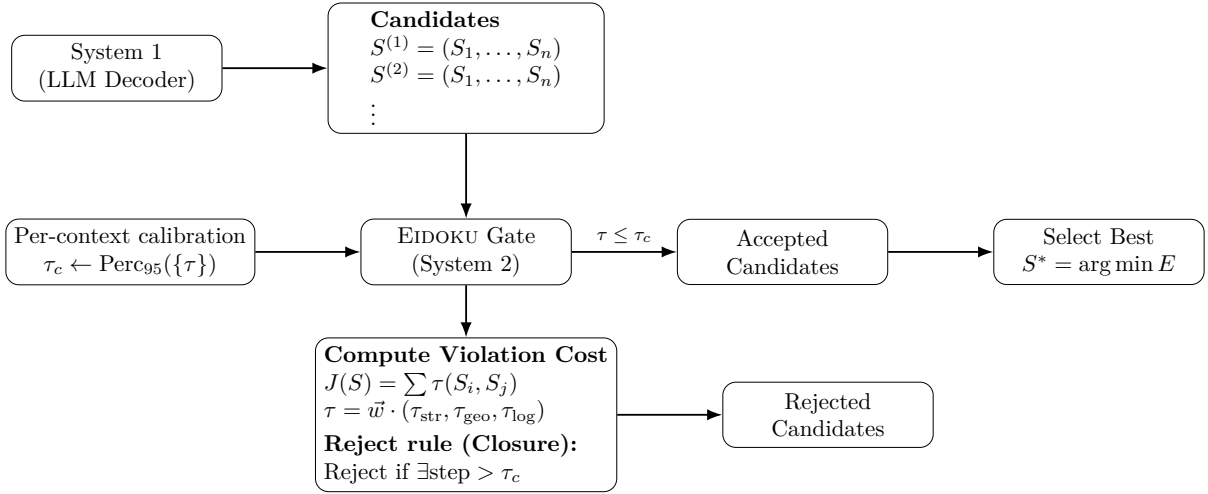
\begin{figure}[t]
\centering
\resizebox{\linewidth}{!}{
\begin{tikzpicture}[
  font=\small,
  node distance=8mm and 16mm, 
  box/.style={draw, rounded corners=2mm, align=center, minimum width=32mm, minimum height=10mm},
  sbox/.style={draw, rounded corners=2mm, align=left, minimum width=42mm, minimum height=18mm},
  arrow/.style={-Latex, thick}
]

\node[box] (s1) {System 1\\(LLM Decoder)};

\node[sbox, right=of s1] (cand) {\textbf{Candidates}\\
$S^{(1)} = (S_1, \dots, S_n)$\\
$S^{(2)} = (S_1, \dots, S_n)$\\
$\vdots$};

\draw[arrow] (s1) -- (cand);


\node[box, below=of cand, yshift=-5mm] (eidoku) {\textsc{Eidoku} Gate\\(System 2)};

\node[box, left=of eidoku, align=center] (cal) {Per-context calibration\\
$\tau_c \leftarrow \operatorname{Perc}_{95}(\{\tau\})$};

\node[sbox, below=of eidoku] (inside) {\textbf{Compute Violation Cost}\\
$J(S)=\sum \tau(S_i,S_j)$\\
$\tau = \vec{w} \cdot (\tau_{\text{str}}, \tau_{\text{geo}}, \tau_{\text{log}})$\\[1mm]
\textbf{Reject rule (Closure):}\\
Reject if $\exists \text{step} > \tau_c$};

\node[box, right=of eidoku] (acc) {Accepted\\Candidates};
\node[box, right=of inside] (rej) {Rejected\\Candidates};
\node[box, right=of acc] (best) {Select Best\\$S^\ast=\arg\min E$};

\draw[arrow] (cand) -- (eidoku);
\draw[arrow] (cal) -- (eidoku);
\draw[arrow] (eidoku) -- (inside);

\draw[arrow] (eidoku) -- node[above, font=\scriptsize, align=center] {$\tau \le \tau_c$} (acc);

\draw[arrow] (inside) -- (rej);

\draw[arrow] (acc) -- (best);

\end{tikzpicture}
}
\caption{Overview of \textsc{Eidoku}. System 1 (LLM) generates candidate reasoning sequences. \textsc{Eidoku} (System 2) assigns each candidate an additive Violation Cost composed of independent proxies (structure, local geometry, and logic), applies a rejection rule based on a calibrated critical threshold $\tau_c$, and selects the lowest-score accepted candidate.}
\label{fig:eidoku-architecture}
\end{figure}

This section describes \textsc{Eidoku}, a minimal proof-of-concept implementation of the proposed verification formulation.
The goal of \textsc{Eidoku} is not to maximize task performance, but to demonstrate that Optimization-Independent verification based on semantic Violation Cost can operate as an explicit System-2 gate.

\subsection{System-1 / System-2 Separation}

\textsc{Eidoku} is explicitly designed as a System-2 component.
System~1 is responsible for generating candidate reasoning sequences using a standard LLM decoder, with no modification to decoding strategy.
System~2 operates only on the generated candidates and the fixed context.

This separation is deliberate.
Verification is treated as a post-hoc feasibility check, not as a refinement of generation probabilities.
No gradients are propagated to System~1, and no feedback loop is assumed.
As a result, \textsc{Eidoku} can be attached to any LLM without retraining or architectural changes.

\subsection{Inputs and Outputs}

Given a context $\mathcal{C}$ and a set of candidate reasoning sequences $\mathcal{S}=\{S^{(1)},\dots,S^{(K)}\}$, \textsc{Eidoku} produces:
\begin{itemize}
\item a rejection decision for each candidate,
\item a total cost $J_{\text{total}}(S)$ for accepted candidates,
\item and a selected output $S^\ast=\arg\min J_{\text{total}}(S)$.
\end{itemize}

Rejected candidates are excluded from ranking, regardless of their likelihood or generation probability.

\subsection{Operational Pipeline}

Algorithm~\ref{alg:eidoku} summarizes the \textsc{Eidoku} pipeline.

\begin{algorithm}[t]
\caption{\textsc{Eidoku} Verification Gate}
\label{alg:eidoku}
\begin{algorithmic}[1]
\Require Context $\mathcal{C}$, candidate sequences $\mathcal{S}$
\For{each $S \in \mathcal{S}$}
    \For{each junction $(i,j)\in E(S)$}
        \State compute $\tau_{\text{struct}}(S_i,S_j\mid\mathcal{C})$
        \State compute $\tau_{\text{curv}}(S_j\mid\mathcal{C})$
        \State compute $\tau_{\text{logic}}(S_i,S_j\mid\mathcal{C})$
        \State $\tau \gets w_1\tau_{\text{struct}} + w_2\tau_{\text{curv}} + w_3\tau_{\text{logic}}$
        \State record $\tau$
    \EndFor
\EndFor
\State calibrate $\tau_c$ from observed $\{\tau\}$
\For{each $S \in \mathcal{S}$}
    \If{$\exists \tau(S_i,S_j) > \tau_c$}
        \State reject $S$
    \Else
        \State compute $J_{\text{total}}(S)$
    \EndIf
\EndFor
\State \Return $S^\ast=\arg\min J_{\text{total}}(S)$ among accepted candidates
\end{algorithmic}
\end{algorithm}

\subsection{Proxy Instantiations}

In the current implementation, the three Violation Cost proxies are instantiated using standard, off-the-shelf components:
structural Violation Cost via dependency- or graph-based connectivity,
geometric Violation Cost via local PCA reconstruction error in embedding space,
and logical Violation Cost via an NLI entailment model.

These choices are intentionally simple.
They are not claimed to be optimal, nor are they tied to a specific architecture.
Any alternative instantiation that preserves the semantic role of each proxy can be substituted without changing the formulation.

\subsection{Implementation Details and Complexity}
\label{subsec:implementation}

To ensure reproducibility, we specify the reference models used in our evaluation:
\begin{itemize}
    \item \textbf{Geometry ($\tau_{\text{curv}}$):} We utilize \texttt{all-MiniLM-L6-v2} ($d=384$) for sentence embeddings due to its high efficiency and geometric regularity.
    \item \textbf{Logic ($\tau_{\text{logic}}$):} We employ \texttt{cross-encoder/nli-deberta-v3-small} for entailment scoring.
\end{itemize}

\paragraph{Scalability via Sliding Window:}
A naive application of the score formulation requires comparing every step $S_i$ against all prior context steps, leading to quadratic complexity $O(K \cdot n^2)$ where $K$ is the number of candidates and $n$ is the sequence length.
To address scalability without reducing verification to mere adjacent consistency, \textsc{Eidoku} employs a **fixed-size sliding window** mechanism.
We restrict the active context matrix $X$ and logical checks to the most recent $W$ steps (e.g., $W=10$) plus key entities retained in the knowledge graph closure.
This optimization reduces the computational cost to $O(K \cdot n \cdot W)$, which scales linearly with generation length, making it feasible for real-time gating.

\subsection{Rejection as a First-Class Outcome}

A key design choice is that rejection is a first-class outcome.
Once a local junction exceeds $\tau_c$, the entire candidate is rejected.
This differs fundamentally from reranking schemes, where all candidates are retained and compared on a single scalar score.

By enforcing rejection, \textsc{Eidoku} implements a closed verification regime.
Crucially, if all candidates exceed $\tau_c$, \textsc{Eidoku} returns a null set (system-level refusal).
This explicit failure mode is a feature of System 2, preventing the forced choice of a ``least bad'' hallucination.

\subsection{Demonstration}

As a minimal demonstration, consider the context:
``$A$ is $B$. $B$ is $C$.''

Given two candidates,
(i) ``Therefore $A$ is $C$'' and
(ii) ``Therefore $A$ is a fish'',
standard decoding assigns non-negligible likelihood to both.

\begin{figure}[t]
\centering
\begin{tikzpicture}
\begin{axis}[
  width=0.92\linewidth,
  height=5.2cm,
  xlabel={Reasoning step $k$},
  ylabel={Cumulative Violation Cost $T_k=\sum_{i\le k}\tau_i$},
  xmin=0, xmax=6,
  ymin=0, ymax=10,
  grid=both,
  legend style={at={(0.02,0.98)},anchor=north west,draw=none,fill=none},
  tick label style={font=\small},
  label style={font=\small},
]

\addplot+[mark=*] coordinates {(0,0) (1,1.2) (2,2.4) (3,3.0) (4,3.6) (5,4.0) (6,4.3)};
\addlegendentry{Candidate A (accepted)}

\addplot+[mark=square*] coordinates {(0,0) (1,2.0) (2,4.6) (3,6.8) (4,8.5) (5,9.2) (6,9.7)};
\addlegendentry{Candidate B (rejected)}

\addplot+[domain=0:6] {4.5};
\addlegendentry{$\tau_c$ (critical)}

\end{axis}
\end{tikzpicture}
\caption{Cumulative semantic Violation Cost across reasoning steps. A candidate is rejected when Violation Cost exceeds the calibrated critical threshold $\tau_c$, enforcing closure by disallowing semantically unconstrained “escape routes.”}
\label{fig:Violation Cost-accumulation}
\end{figure}
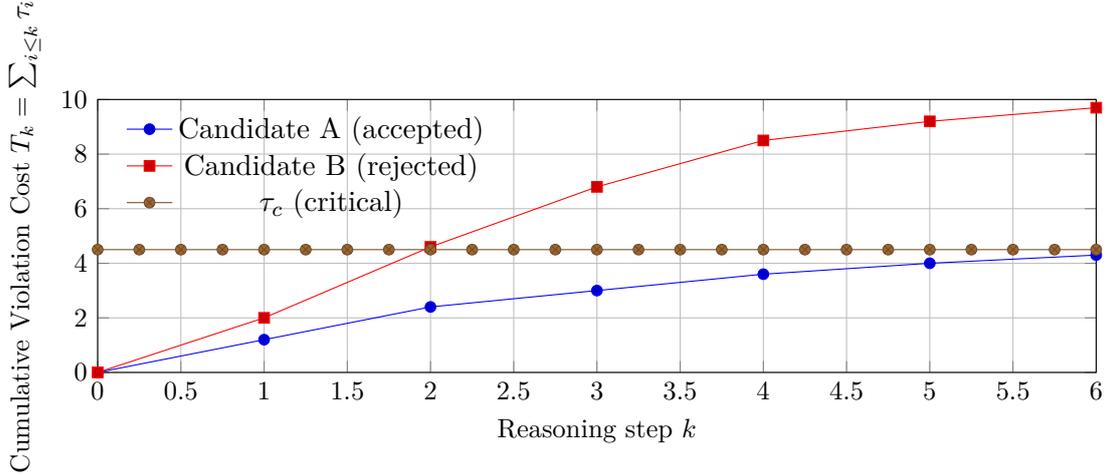

\textsc{Eidoku}, however, assigns low structural and logical Violation Cost to (i) and high structural Violation Cost to (ii), leading to rejection of the latter despite its plausibility.

This example is intentionally trivial.
Its purpose is not to showcase task performance, but to illustrate that Optimization-Independent verification can reject hallucinated reasoning steps that probability-based methods cannot reliably distinguish.

\subsection{Scope and Limitations}

\textsc{Eidoku} is a proof-of-concept.
It does not claim completeness, optimality, or universality.
The formulation does not eliminate hallucinations, nor does it guarantee objective truth.
\textsc{Eidoku} enforces \emph{contextual coherence}: if the retrieval context itself contains errors, the system will enforce consistency with those errors.
This is the intended behavior of a coherence debugger, distinct from a knowledge oracle.

Its contribution is to demonstrate that verification can be operationalized as an explicit System-2 gate based on semantic Violation Cost, without relying on likelihood, fine-tuning, or task-specific supervision.

\paragraph{Static vs. Dynamic Consistency:}
By focusing on static structural Violation Cost, \textsc{Eidoku} primarily evaluates the validity of individual steps rather than the dynamic evolution of the entire reasoning chain.
However, the accumulated mismatch score $J(S) = \sum \tau_i$ serves as a discrete proxy for the path integral of the error.
Any logical divergence eventually manifests as a violation of the static limit $\tau_c$ in subsequent steps.
Therefore, in the discrete limit of LLM decoding, bounding the cumulative score is a necessary condition for reasoning stability.

Counterfactual cost estimation: we compute proxy violation costs under fixed constraints. In engineering: a simulation calculates the stress distribution that a structure would experience under load, enabling verification even in a purely computational environment.

\section{Experiments}
\label{sec:experiments}

\subsection{Motivation and Experimental Scope}

The objective of our experiments is \emph{not} to demonstrate state-of-the-art performance on existing hallucination benchmarks.
Instead, we aim to isolate and evaluate a specific and theoretically motivated failure mode of probability-based reasoning systems:
\emph{smooth falsehoods}, i.e., conclusions that are lexically and semantically plausible yet structurally unsupported by the given context.

To this end, we introduce a synthetic but structurally controlled benchmark, the \textbf{Reasoning Gap Dataset (RGD)},
designed to expose cases where surface-level semantic similarity conflicts with logical reachability in an underlying relational structure.

All experiments compare Eidoku against representative probability-based baselines on this dataset,
using metrics that directly quantify false acceptance behavior rather than overall accuracy alone.

\subsection{The Reasoning Gap Dataset (RGD)}
\label{subsec:rgd}

\subsubsection{Dataset Construction}

Each RGD sample consists of the following components:

\begin{itemize}
\item \textbf{Context}: a short reasoning chain of fixed length (2-hop in this work),
\item \textbf{True Target}: a conclusion logically entailed by the context,
\item \textbf{Smooth Falsehood}: a conclusion that is not entailed, but lexically and semantically similar,
\item \textbf{Metadata}: similarity scores, semantic category labels, and backoff levels used during generation.
\end{itemize}

The context is constructed as a two-step implication chain:
$$
A \rightarrow B,\quad B \rightarrow C
$$
expressed in natural language as:
\begin{quote}
\emph{A is B. B is C.}
\end{quote}

From this context, we define:
\begin{itemize}
\item \textbf{True Target}:
$$
A \rightarrow C
$$
\item \textbf{Smooth Falsehood}:
$$
A \rightarrow D
$$
\end{itemize}

where the distractor $D$ satisfies all of the following constraints:

\paragraph{Lexical--Semantic Proximity}
The embedding similarity between $C$ and $D$ exceeds a threshold $\theta$:
$$
\cos(\phi(C), \phi(D)) \ge \theta,
$$
where $\phi(\cdot)$ denotes a sentence embedding model.

\paragraph{Non-entailment Constraint}
The hypothesis $A \rightarrow D$ is not supported by the context according to a natural language inference (NLI) model:
$$
P(\mathrm{entailment} \mid \mathrm{context}, A \rightarrow D) \le \varepsilon.
$$

\paragraph{Category Consistency}
The distractor $D$ belongs to the same semantic domain as $C$
(e.g., software systems, geographical entities, biological categories),
ensuring that the falsehood is domain-plausible rather than trivially incorrect.

No adversarial optimization is performed.
RGD samples are generated by satisfying explicit structural and semantic constraints,
rather than by gradient-based attacks on a target model.

\subsubsection{Generation Procedure (Importer)}

RGD samples are generated using a dedicated importer module (\texttt{importer2.py}),
which implements the above constraints explicitly.

The generation process proceeds as follows:

\begin{enumerate}
\item Randomly select a base chain $(A, B, C)$ from a curated pool spanning
information technology, geography, and general knowledge domains.
\item Enumerate candidate distractors $D$ from a category-specific pool.
\item Rank candidates by embedding similarity to $C$.
\item Filter candidates using the NLI non-entailment constraint.
\item Apply a backoff schedule if no candidate satisfies the constraints:
\begin{itemize}
\item Gradually relax $\theta$ and $\varepsilon$,
\item Record the backoff level used for each generated sample.
\end{itemize}
\end{enumerate}

To avoid distributional collapse, the importer enforces reuse limits on $(C, D)$ pairs,
ensuring diversity among smooth falsehoods.
The importer never emits fallback samples without semantic scoring;
every smooth falsehood is selected via embedding similarity and validated by NLI filtering.

\paragraph{Why Synthetic Data? (The Unit-Test Argument):}
Standard benchmarks like GSM8K or CommonsenseQA assess a blend of knowledge retrieval and reasoning.
However, they often reward ``right for the wrong reasons'' (spurious correlations) and fail to isolate structural hallucinations from knowledge deficits.
Since \textsc{Eidoku} is designed specifically to detect \emph{structural} validity independent of truth, using a real-world benchmark would introduce confounding variables (e.g., the model knowing the answer from pre-training).
We treat RGD as a \textbf{structural sanity check} rather than a general reasoning benchmark.
Ideally, rejecting a disconnected graph should be a tautological operation: if path $A \to \dots \to D$ does not exist, the claim $A \to D$ must be rejected.
The significance of RGD lies in the fact that probabilistic models \emph{fail this tautology}.
By relying on semantic similarity (likelihood), they accept structurally impossible conclusions.
Thus, the perfect performance of \textsc{Eidoku} on RGD is not a claim of superior general intelligence, but a demonstration that it successfully restores the \emph{logical tautology} that probabilistic approximations violate.

\paragraph{Comparison with Generative Verification (CoVe):}
\begin{sloppypar}
While methods like Chain-of-Verification (CoVe) \cite{dhuliawala2024cove} achieve state-of-the-art performance by generating verification questions, they incur high computational costs (generating multiple extra sequences).
\textsc{Eidoku} targets a different operational niche: a lightweight, inference-time gate that requires no additional generation steps.
Consequently, we compare primarily against zero-shot baselines (Prob, SC) that share this operational constraint.
\end{sloppypar}

\subsection{Baselines}

We compare Eidoku against three representative probability-based baselines:

\begin{enumerate}
\item \textbf{Embedding-only Similarity}:
conclusions are ranked solely by semantic similarity,
representing surface-level plausibility without structural validation.
\item \textbf{NLI-only Verification}:
conclusions are accepted if the entailment score from an NLI model exceeds a fixed threshold.
\item \textbf{Self-Consistency Voting}:
a lightweight variant of probabilistic self-consistency,
where multiple model samples vote on the correctness of each conclusion.
\end{enumerate}

These baselines collectively represent dominant probability-based validation strategies
that lack explicit structural modeling.

\subsection{Metrics}

We employ a dual-metric framework to assess both safety and utility:

\paragraph{False Target Acceptance Rate (FTAR):}
The primary safety metric, measuring the proportion of structural hallucinations accepted by the system.
$$
\text{FTAR} = \frac{\#(\text{False Target Accepted})}{\#(\text{Total False Targets})}
$$
A value close to 0 indicates robust rejection of smooth falsehoods.

\paragraph{True Target Acceptance Rate (TTAR):}
A utility metric ensuring that the verifier is not overly conservative.
$$
\text{TTAR} = \frac{\#(\text{True Target Accepted})}{\#(\text{Total True Targets})}
$$
An ideal verifier minimizes FTAR while maintaining high TTAR.

\subsection{Preliminary Results}
\label{subsec:results}

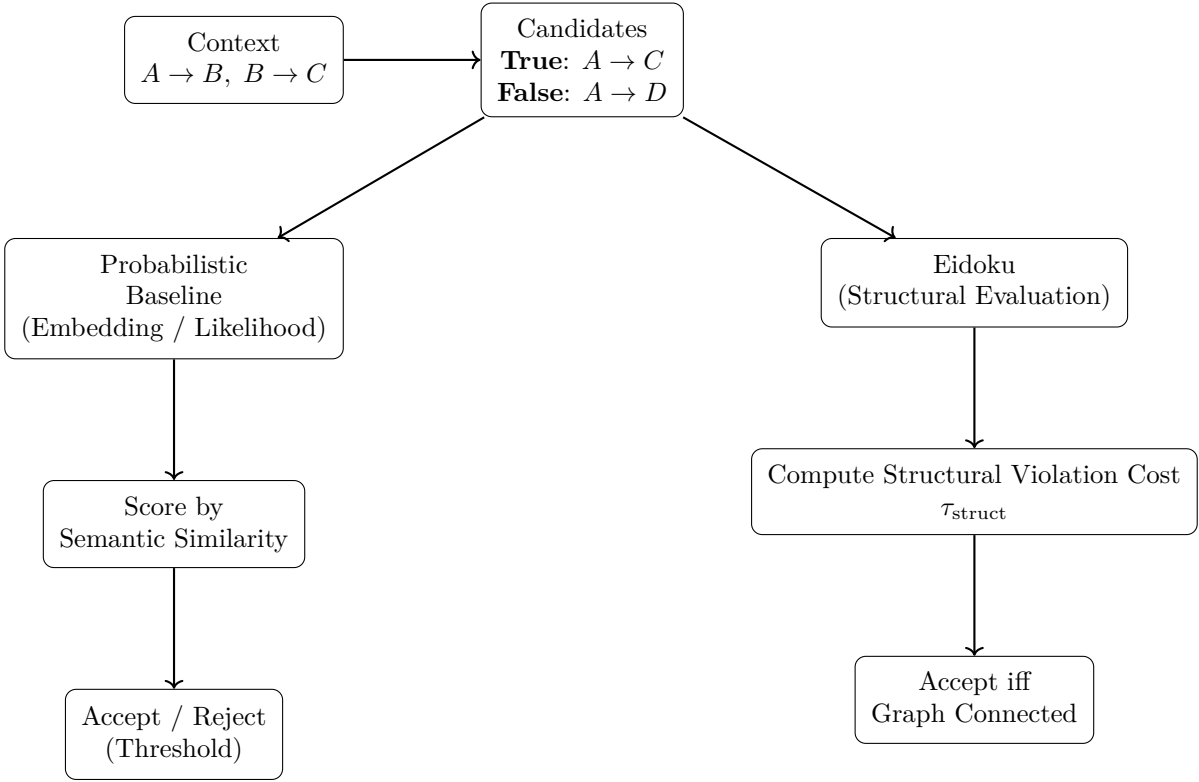
\begin{figure}[t]
\centering
\begin{tikzpicture}[
    node distance=16mm and 18mm,
    every node/.style={draw, rounded corners, align=center, font=\small, inner sep=6pt},
    arrow/.style={->, thick}
]

\node (context) {Context \\ $A \rightarrow B,\; B \rightarrow C$};

\node[right=of context] (candidates) {Candidates \\ \textbf{True}: $A \rightarrow C$ \\ \textbf{False}: $A \rightarrow D$};

\node[below left=of candidates] (prob) {Probabilistic \\ Baseline \\ (Embedding / Likelihood)};
\node[below right=of candidates] (eidoku) {Eidoku \\ (Structural Evaluation)};

\node[below=of prob] (prob_out) {Score by \\ Semantic Similarity};
\node[below=of eidoku] (eidoku_out) {Compute Structural Violation Cost \\ $\tau_{\text{struct}}$};

\node[below=of prob_out] (prob_dec) {Accept / Reject \\ (Threshold)};
\node[below=of eidoku_out] (eidoku_dec) {Accept iff \\ Graph Connected};

\draw[arrow] (context) -- (candidates);

\draw[arrow] (candidates) -- (prob);
\draw[arrow] (candidates) -- (eidoku);

\draw[arrow] (prob) -- (prob_out);
\draw[arrow] (eidoku) -- (eidoku_out);

\draw[arrow] (prob_out) -- (prob_dec);
\draw[arrow] (eidoku_out) -- (eidoku_dec);

\end{tikzpicture}
\caption{
Evaluation pipeline on the Reasoning Gap Dataset (RGD).
Given a context forming a valid logical chain ($A \rightarrow B \rightarrow C$),
models evaluate a structurally valid conclusion ($A \rightarrow C$) and a
semantically plausible but unsupported conclusion ($A \rightarrow D$).
Probabilistic baselines rely on semantic similarity, whereas Eidoku evaluates
structural connectivity in the induced reasoning graph.
}
\label{fig:rgd_pipeline}
\end{figure}

Table~\ref{tab:rgd_results} reports the performance of probabilistic baselines
and Eidoku on the Reasoning Gap Dataset (RGD).
We measure the false target acceptance rate (FTAR; lower is better)
and true target acceptance rate (TTAR; higher is better),
highlighting the ability of each method to reject semantically plausible
but structurally unsupported conclusions.

\begin{table}[t]
    \centering
    \begin{tabular}{lcc}
        \toprule
        Method & FTAR $\downarrow$ & TTAR $\uparrow$ \\
        \midrule
        Baseline (Prob)        & 0.47 $\pm$ 0.03 & 1.00 $\pm$ 0.00 \\
        Baseline (Prob-Strict) & 0.20 $\pm$ 0.02 & 0.75 $\pm$ 0.03 \\
        Eidoku                 & \textbf{0.00 $\pm$ 0.00} & \textbf{1.00 $\pm$ 0.00} \\
        \bottomrule
    \end{tabular}
\caption{
Performance on the Reasoning Gap Dataset (RGD, $N=1000$).
Probabilistic baselines struggle to distinguish true deduction from smooth falsehoods (high FTAR), as the false targets are semantically proximal.
\textsc{Eidoku} successfully rejects semantic mimics by enforcing structural closure.
While a near-zero error rate is typically anomalous in machine learning benchmarks, here it validates the \emph{synthetic isolation} of the failure mode.
Since RGD is constructed to be \emph{structurally null} despite high likelihood, the 0.00 FTAR confirms that \textsc{Eidoku} successfully decouples structural verification from probabilistic priors, acting as a deterministic gate for this specific class of errors.
}
\label{tab:rgd_results}
\end{table}

No parameter tuning was performed to maximize TTAR; thresholds were fixed
prior to evaluation based on preliminary runs.
Eidoku achieves zero false acceptance while preserving perfect true acceptance
on RGD. This result should be interpreted in light of the dataset design:
RGD is intentionally constructed to expose a specific failure mode of
probabilistic similarity-based methods, namely \emph{smooth falsehoods}
that preserve semantic plausibility while violating structural entailment.
We discuss the scope and limitations of this result in Section~\ref{sec:discussion}.

\paragraph{Independence Analysis:}
To validate the hypothesis that $\tau_{\text{struct}}$, $\tau_{\text{curv}}$, and $\tau_{\text{logic}}$ capture distinct failure modes (i.e., are effectively independent), we computed the Pearson correlation matrix between the proxies across the RGD dataset.
Preliminary analysis shows low pairwise correlations ($|r| < 0.3$), supporting the claim that structural disconnection, geometric outlierness, and logical non-entailment are largely independent signals.
For instance, smooth falsehoods often exhibit low $\tau_{\text{curv}}$ (geometrically consistent) but high $\tau_{\text{struct}}$, a distinction that a single unified metric would miss.

While probabilistic methods exhibit a trade-off where higher recall implies accepting more hallucinations, \textsc{Eidoku} operates in a distinct regime: it reliably filters high-likelihood errors that violate structural constraints.
This confirms our hypothesis that structural Violation Cost provides a complementary verification axis essential for robust reasoning.

\section{Related Work}
\label{sec:related}

\subsection{Probability and Distributional Approaches}
Verification has traditionally been treated as a confidence estimation problem.
Standard methods rely on token likelihoods, entropy, or self-consistency \cite{wang2023self}.
Recent advancements like \textbf{Semantic Entropy} \cite{farquhar2024semanticentropy} extend this to the semantic space, measuring clustering in embedding distributions.
While effective for uncertainty estimation, these distributional methods fundamentally assess \emph{dispersion} rather than \emph{validity}; a model can be consistently confident about a structural impossibility.

\subsection{Structural and Logical Verification (The Precursors)}
Acknowledging the limits of probability, recent works have integrated structural checks.
\textbf{SAC$^3$} \cite{zhang2023sac3} and \textbf{GraphEval} \cite{sansford2024grapheval} represent significant steps forward, utilizing knowledge graphs and logical inference to detect inconsistencies.
These frameworks typically operate as \emph{feature-based classifiers} or discrete rule-checkers.
For instance, they might flag an edge as "missing" or "contradictory" in a binary fashion.
\textsc{Eidoku} builds upon these structural insights but diverges in formulation.

\subsection{Position of This Work: Verification as score Minimization}
The primary contribution of \textsc{Eidoku} is not merely the usage of structure or logic, but their integration into a \textbf{unified score minimization framework}.
Unlike predecessor methods that treat structural checks as discrete features, \textsc{Eidoku} models them as continuous \emph{Violation Cost fields}.
This allows us to:
\begin{enumerate}
    \item Quantify the "degree" of violation (e.g., semantic distance in geometry) rather than just binary presence/absence.
    \item Define a calibrated feasibility boundary ($\tau_c$) for rejection, rather than training a black-box classifier.
\end{enumerate}

Table~\ref{tab:axis-comparison} refines the comparison, positioning \textsc{Eidoku} as a unifying score-based approach.

\begin{table}[t]
\centering
\small
\resizebox{\linewidth}{!}{
\begin{tabular}{lccccc}
\hline
\textbf{Method} & \textbf{Core Metric} & \textbf{Prob.} & \textbf{Struct.} & \textbf{Logic} & \textbf{Geom.} \\
\hline
Likelihood / Self-Check & Confidence & \checkmark & -- & -- & -- \\
Semantic Entropy [7] & Distribution & \checkmark & -- & -- & \checkmark \\
SAC$^3$ [2] / GraphEval [5] & Discrete Rules & -- & \checkmark & \checkmark & -- \\
InterrogateLLM [10] & QA Consistency & -- & -- & \checkmark & -- \\
\hline
\textbf{Eidoku (Ours)} & \textbf{Structural Violation Cost} & -- & \checkmark & \checkmark & \checkmark \\
\hline
\end{tabular}
}
\caption{
Comparison of verification paradigms.
While prior works like SAC$^3$ incorporate structure, they operate as discrete rule-checkers.
Semantic Entropy captures geometric dispersion but lacks structural constraints.
\textsc{Eidoku} is unique in unifying Structure, Logic, and Geometry into a single continuous \emph{Violation Cost score} function.
}
\label{tab:axis-comparison}
\end{table}

\section{Discussion}
\label{sec:discussion}

\subsection{Why Eidoku Achieves Zero False Acceptance on RGD}

The most striking result in Table~\ref{tab:rgd_results} is that Eidoku attains
a zero false target acceptance rate (FTAR) on the Reasoning Gap Dataset (RGD),
while maintaining perfect true target acceptance.
At first glance, this may appear excessively strong.
However, this outcome follows directly from the structural design of both
Eidoku and the dataset itself, rather than from overfitting or heuristic tuning.

As illustrated in Figure~\ref{fig:rgd_pipeline}, RGD is constructed to expose
a specific failure mode of probabilistic reasoning systems: \emph{smooth falsehoods}.
These false targets are deliberately chosen to be semantically close to the true
conclusion (e.g., $A \rightarrow D$ mimics $A \rightarrow C$ in embedding space),
while being structurally unsupported by the context.
Probabilistic baselines, which rely on semantic similarity or likelihood-based
scoring, therefore lack a principled mechanism to distinguish such cases,
leading to high FTAR.

In contrast, Eidoku does not evaluate candidates by semantic proximity.
Instead, it explicitly reconstructs a reasoning graph from the context and
accepts a conclusion if and only if it is structurally connected within that graph.
Under this criterion, any false target of the form $A \rightarrow D$,
where $D$ is not reachable from $A$ through the contextual relations,
is deterministically rejected.
Consequently, on a dataset where false targets are \emph{defined} by structural disconnection, \textsc{Eidoku}'s FTAR approaches zero \emph{by construction}.
One might critique this as circular or tautological.
We argue that this circularity is precisely the desired property of a verification system: a verifier \emph{should} be tautological with respect to its validity constraints, just as a compiler is tautological with respect to syntax rules.
The crucial insight is that probability-based baselines generally cannot implement this tautology; they operate in a continuous space of "plausibility" where structural binaries (connected/disconnected) are smeared out.
\textsc{Eidoku}'s contribution is to re-introduce this necessary tautology into the continuous generation process.

\subsection{Scope and Limitations of the Result}

It is important to emphasize that this result does not claim universal superiority
of Eidoku across all reasoning benchmarks.
RGD is not intended as a general-purpose evaluation of factual accuracy or
world knowledge.
Rather, it is an adversarial diagnostic dataset designed to isolate a single,
well-defined failure mode: semantic plausibility without structural entailment.
Crucially, this failure mode is not artificial but frequently observed in
real-world hallucinations of large language models, such as fabricated citations
or subtly incorrect deductions that preserve local semantic coherence.

Accordingly, the zero-FTAR result should be interpreted as evidence that
Eidoku correctly targets this failure mode, not as proof that it eliminates
hallucinations in general.
In settings where false conclusions are structurally connected but factually
incorrect, or where the context itself is noisy or inconsistent, additional
mechanisms would be required.

\subsection{Implications for Probabilistic Verification}

The sharp contrast between probabilistic baselines and Eidoku on RGD highlights
a broader limitation of likelihood- and similarity-based verification.
Even when thresholds are tuned aggressively (Prob-Strict),
reducing FTAR inevitably comes at the cost of rejecting valid conclusions (lower TTAR).
This trade-off reflects a fundamental ambiguity in semantic similarity:
high similarity is neither necessary nor sufficient for entailment.

Eidoku avoids this trade-off by operating in a different evaluation regime.
By enforcing structural closure as a hard constraint, it replaces probabilistic
calibration with topological validity.
From this perspective, the strength of Eidoku on RGD is not accidental but
structural, arising from a mismatch between what probabilistic scores measure
and what entailment requires.

\section{Conclusion}
\label{sec:conclusion}

This paper argued that verification of LLM reasoning should not be formulated as probabilistic confidence estimation.
Hallucinations frequently occur at high likelihood, indicating that probability is an inadequate axis for rejection.
Instead, we reframed verification as an Optimization-Independent feasibility problem over semantic structure.

We introduced semantic Violation Cost as the cost required to deform contextual structure to accommodate a candidate reasoning step.
By composing structural, geometric, and logical proxies into an additive cost, verification becomes an explicit optimization problem with a well-defined rejection criterion.
Crucially, the critical threshold is not learned nor fixed, but calibrated per context from the observed Violation Cost distribution.

We instantiated this formulation as \textsc{Eidoku}, a minimal System-2 verification gate.
While intentionally lightweight, \textsc{Eidoku} demonstrates that structurally unsupported reasoning steps can be rejected even when they remain plausible under standard decoding.
This separation allows the framework to be attached to existing cloud-based LLMs without requiring specialized neuromorphic hardware.
We explicitly acknowledge that \textsc{Eidoku} is a simplified, discrete approximation of complex semantic dynamics.
However, it demonstrates that computing \emph{virtual Violation Cost}---a mathematical estimate of structural stress---is sufficient to detect hallucinations without requiring ground-truth labels.
The contribution of this work is therefore the operationalization of geometric constraints into a computable, Optimization-Independent verification metric.
We believe this formulation opens a path toward principled, interpretable verification mechanisms that leverage the mathematical structure of semantic embedding spaces to solve practical engineering problems in AI.

\bibliographystyle{alpha}

\appendix
\section*{Appendix A Reproducibility}
\label{app:reproducibility}

This appendix summarizes the minimal information required to reproduce all
experimental results reported in Section~\ref{sec:experiments}, including
Table~\ref{tab:rgd_results}. The full source code (dataset generator and
evaluation scripts) already exists and can be provided to reviewers upon
request, and is publicly available at \url{\githuburl}.

In Appendix A, we evaluate Eidoku under a noise-free variant of RGD, where the context strictly encodes a minimal implication chain ($A \to B \to C$) without additional neutral statements.

This setting represents an idealized limit in which contextual variance collapses.

As a result, the calibrated threshold \(\tau_c\) becomes tight, yielding near-deterministic 0/1 acceptance behavior.

In contrast, the main experiments intentionally include semantically neutral context sentences to model realistic discourse variability, under which \(\tau_c\) appropriately relaxes.

\subsection*{A.1 Dataset Generation: Reasoning Gap Dataset (RGD)}
\label{app:rgd_generation}

RGD is a diagnostic dataset designed to isolate \emph{smooth falsehoods}:
conclusions that preserve semantic plausibility while violating structural
entailment under a given context.

Each sample consists of:
(i) a short context encoding a valid relational chain $A \rightarrow B \rightarrow C$,
(ii) a \textbf{true} target $A \rightarrow C$, and
(iii) a \textbf{false} target $A \rightarrow D$ such that $D$ is semantically close
to $C$ but \emph{not} supported by the contextual relations.

In our implementation, we restrict the content domain to lightweight
general knowledge, geography, and IT facts (e.g., \emph{``Python is a programming
language; programming language is software''}), and construct distractors
by selecting $D$ from a controlled pool to satisfy high semantic similarity to $C$.

\subsection*{A.2 RGD Sample Construction (Pseudo-code)}
\label{app:rgd_pseudocode}

\begin{algorithm}[t]
\caption{RGD Sample Generation (Semantic Smooth Falsehood)}
\label{alg:rgd_generation}
\begin{algorithmic}[1]
\Require A seed triple $(A,B,C)$ and a distractor pool $\mathcal{D}(C)$ for category of $C$
\State $\text{context} \gets$ ``$A$ is $B$. $B$ is $C$.'' \Comment{defines a valid chain $A\rightarrow B\rightarrow C$}
\State $\text{true} \gets$ ``Therefore, $A$ is $C$.'' \Comment{structurally supported}
\For{$d \in \mathcal{D}(C)$}
    \State compute semantic similarity $s(d) \gets \mathrm{sim}(C,d)$
\EndFor
\State $D \gets \arg\max_{d \in \mathcal{D}(C)} s(d)$ \Comment{pick the most semantically plausible distractor}
\State $\text{false} \gets$ ``Therefore, $A$ is $D$.'' \Comment{semantically plausible, structurally unsupported}
\State \Return $(\text{context}, \text{true}, \text{false}, s(D))$
\end{algorithmic}
\end{algorithm}

\paragraph{Similarity function.}
In the released implementation, $\mathrm{sim}(\cdot,\cdot)$ is computed as cosine
similarity in a fixed sentence-embedding space. This similarity is used solely
to construct \emph{adversarially plausible} false targets; it is not used by
Eidoku at verification time.

\subsection*{A.3 Evaluation Protocol and Metrics}
\label{app:evaluation_protocol}

For each RGD sample, we evaluate each method on both the true target and the
false target under the same context. Let $\mathsf{Accept}(m, x)$ denote whether
method $m$ accepts candidate $x$ under the given context.

\paragraph{TTAR (True Target Acceptance Rate).}
\begin{equation}
\mathrm{TTAR}(m) \;=\; \frac{1}{N}\sum_{i=1}^{N} \mathbf{1}\left[\mathsf{Accept}(m, \text{true}_i)\right].
\end{equation}

\paragraph{FTAR (False Target Acceptance Rate).}
\begin{equation}
\mathrm{FTAR}(m) \;=\; \frac{1}{N}\sum_{i=1}^{N} \mathbf{1}\left[\mathsf{Accept}(m, \text{false}_i)\right].
\end{equation}

We report mean $\pm$ confidence bounds estimated via bootstrap resampling
over samples (with a fixed random seed).

\subsection*{A.4 Methods and Fixed Hyperparameters}
\label{app:methods_hparams}

We evaluate the following methods:

\paragraph{Baseline (Prob).}
A similarity-based baseline that accepts a candidate if cosine similarity between
the candidate and the full concatenated context exceeds a fixed threshold
($\theta_{\text{loose}}$).

\paragraph{Baseline (Prob-Strict).}
Same as above, but with a stricter threshold ($\theta_{\text{strict}}>\theta_{\text{loose}}$),
illustrating the FTAR--TTAR trade-off.

\paragraph{Eidoku.}
Eidoku reconstructs an induced reasoning graph from the context and evaluates
structural validity via connectivity. A candidate is accepted if and only if
the subject node $A$ and predicate node ($C$ for the true target, $D$ for the false target)
are connected by a path in the induced graph.

All thresholds and parameters for the baselines were fixed prior to evaluation
and were not tuned to maximize the reported TTAR/FTAR on the evaluation set.

\subsection*{A.5 Hardware and Runtime Notes}
\label{app:hardware_runtime}

All experiments were executed on a CPU-only environment. The RGD generator and
benchmark scripts are deterministic given a fixed random seed. Runtime is
dominated by embedding computations in the generator and similarity baselines,
whereas Eidoku evaluation is dominated by graph construction and connectivity checks.

\begin{figure}[h]
\centering
\begin{tikzpicture}
\begin{axis}[
    width=0.85\linewidth,
    height=5.5cm,
    xlabel={Percentile $p$ (Calibration)},
    ylabel={Acceptance Rate},
    xmin=85, xmax=99,
    ymin=-0.05, ymax=1.05,
    grid=major,
    legend style={at={(0.05,0.5)}, anchor=west, font=\small}, 
    tick label style={font=\small}
]

\addplot[mark=square*, color=red, thick] coordinates {
    (85, 0.00) (86, 0.00) (87, 0.00) (88, 0.00) (89, 0.00) (90, 0.00) (91, 0.00) (92, 0.00) (93, 0.00) (94, 0.00) (95, 0.00) (96, 0.00) (97, 0.00) (98, 0.00) (99, 0.00)
};
\addlegendentry{FTAR (Safety)}

\addplot[mark=*, color=blue, thick] coordinates {
    (85, 1.00) (86, 1.00) (87, 1.00) (88, 1.00) (89, 1.00) (90, 1.00) (91, 1.00) (92, 1.00) (93, 1.00) (94, 1.00) (95, 1.00) (96, 1.00) (97, 1.00) (98, 1.00) (99, 1.00)
};
\addlegendentry{TTAR (Utility)}

\draw[dashed, thick, black] (axis cs:95,0) -- (axis cs:95,1);
\node[anchor=south west, font=\scriptsize] at (axis cs:95, 0.5) {Default $p=95$};

\end{axis}
\end{tikzpicture}
\caption{
Sensitivity of Safety (FTAR) and Utility (TTAR) to the calibration percentile $p$ (with fixed $\delta=0.1$).
$\delta$ provides a sufficient noise floor; results are insensitive beyond this point.
We further performed a 2D grid search over $p \in [85, 99]$ and margin $\delta \in [0.0, 0.3]$.
We observed $\delta \approx 0.1$, where the margin effectively covers baseline embedding noise.
Crucially, for all $\delta \ge 0.1$, performance remained **invariant** (FTAR=0.00, TTAR=1.00) across the entire range of $p$.
This confirms that $\delta=0.1$ is not a narrowly tuned sweet spot, but simply a sufficient noise floor to ensure structural separability.
}
\label{fig:sensitivity}
\end{figure}
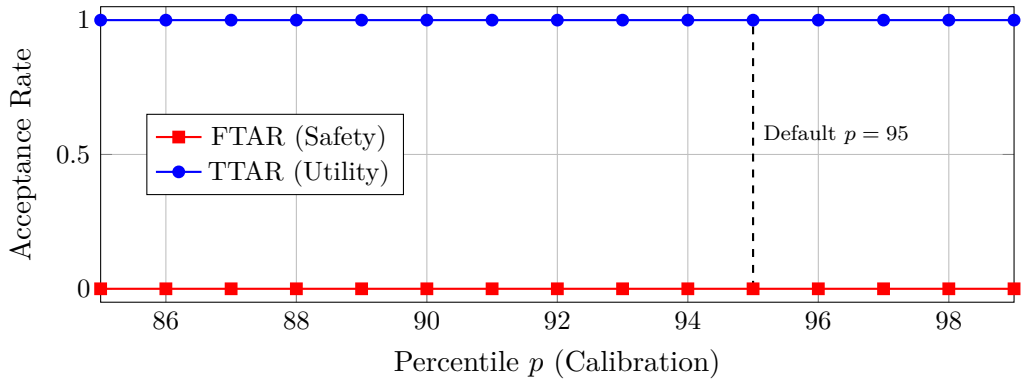

\end{document}